\documentclass[sigconf]{acmart}
\pdfoutput=1
\renewcommand\footnotetextcopyrightpermission[1]{} 

\usepackage{multirow}
\usepackage{fancyhdr}
\AtBeginDocument{%
  \providecommand\BibTeX{{%
    \normalfont B\kern-0.5em{\scshape i\kern-0.25em b}\kern-0.8em\TeX}}}

\acmConference[MM '22]{Proceedings of the 30th ACM International Conference on
Multimedia}{October 10--14, 2022}{Lisboa, Portugul}
\acmBooktitle{Proceedings of the 30th ACM International Conference on Multimedia
(MM '22), October 10--14, 2022, Lisboa, Portugal}
\acmPrice{15.00}
\acmISBN{978-1-4503-9203-7/22/10}

\begin{document}

\title{Incremental Few-Shot Semantic Segmentation via Embedding Adaptive-Update and Hyper-class Representation}
\renewcommand{\shorttitle}{Incremental Few-Shot Semantic Segmentation}

\author{Guangchen Shi}
\email{shiguangchen@hhu.edu.cn}
\affiliation{
  \department{Collage of Computer and Information}
  \institution{Hohai University}
  \city{Nanjing}
  \country{China} 
}

\author{Yirui Wu}
\authornote{corresponding author.}
\email{wuyirui@hhu.edu.cn}
\affiliation{%
  \department{Collage of Computer and Information}
  \institution{Hohai University}
  \city{Nanjing}
  \country{China} 
}

\author{Jun Liu}
\email{junliu@sutd.edu.sg}
\affiliation{%
  \department{Information Systems Technology and Design Pillar}
  \institution{Singapore University of Technology and Design}
  \country{Singapore} 
}

\author{Shaohua Wan}
\email{shaohua.wan@uestc.edu.cn}
\affiliation{%
 \department{Shenzhen Institute for Advanced Study}
  \institution{University of Electronic Science and Technology of China}
  \city{Shenzhen}
  \country{China} 
 }

\author{Wenhai Wang}
\email{wangwenhai@pjlab.org.cn}
\affiliation{%
  \institution{Shanghai AI Laboratory}
  \city{Shanghai}
  \country{China}
 }

\author{Tong Lu}
\email{lutong@nju.edu.cn}
 \affiliation{%
  \department{National Key Lab for Novel Software Technology}
  \institution{Nanjing University}
  \city{Nanjing}
  \country{China} 
}

\renewcommand{\shortauthors}{Guangchen Shi et al.}

\begin{abstract}
  Incremental few-shot semantic segmentation (IFSS) targets at incrementally expanding model's capacity to segment new class of images supervised by only a few samples. However, features learned on old classes could significantly drift, causing catastrophic forgetting. Moreover, few samples for pixel-level segmentation on new classes lead to notorious overfitting issues in each learning session.
  In this paper, we explicitly represent class-based knowledge for semantic segmentation as a category embedding and a hyper-class embedding, where the former describes exclusive semantical properties, and the latter expresses hyper-class knowledge as class-shared semantic properties. 
  Aiming to solve IFSS problems, we present EHNet, i.e., Embedding adaptive-update and Hyper-class representation Network from two aspects.
  First, we propose an embedding adaptive-update strategy to avoid feature drift, which maintains old knowledge by hyper-class representation, and adaptively update category embeddings with a class-attention scheme to involve new classes learned in individual sessions.
  Second, to resist overfitting issues caused by few training samples, a hyper-class embedding is learned by clustering all category embeddings for initialization and aligned with category embedding of the new class for enhancement, where learned knowledge assists to learn new knowledge, thus alleviating performance dependence on training data scale.
  Significantly, these two designs provide representation capability for classes with sufficient semantics and limited biases, enabling to perform segmentation tasks requiring high semantic dependence.
  Experiments on PASCAL-5$^i$ and COCO datasets show that EHNet achieves new state-of-the-art performance with remarkable advantages.
\end{abstract}

 

\maketitle

\section{Introduction}

\begin{figure}[t]
  \centering
  \includegraphics[width=\linewidth]{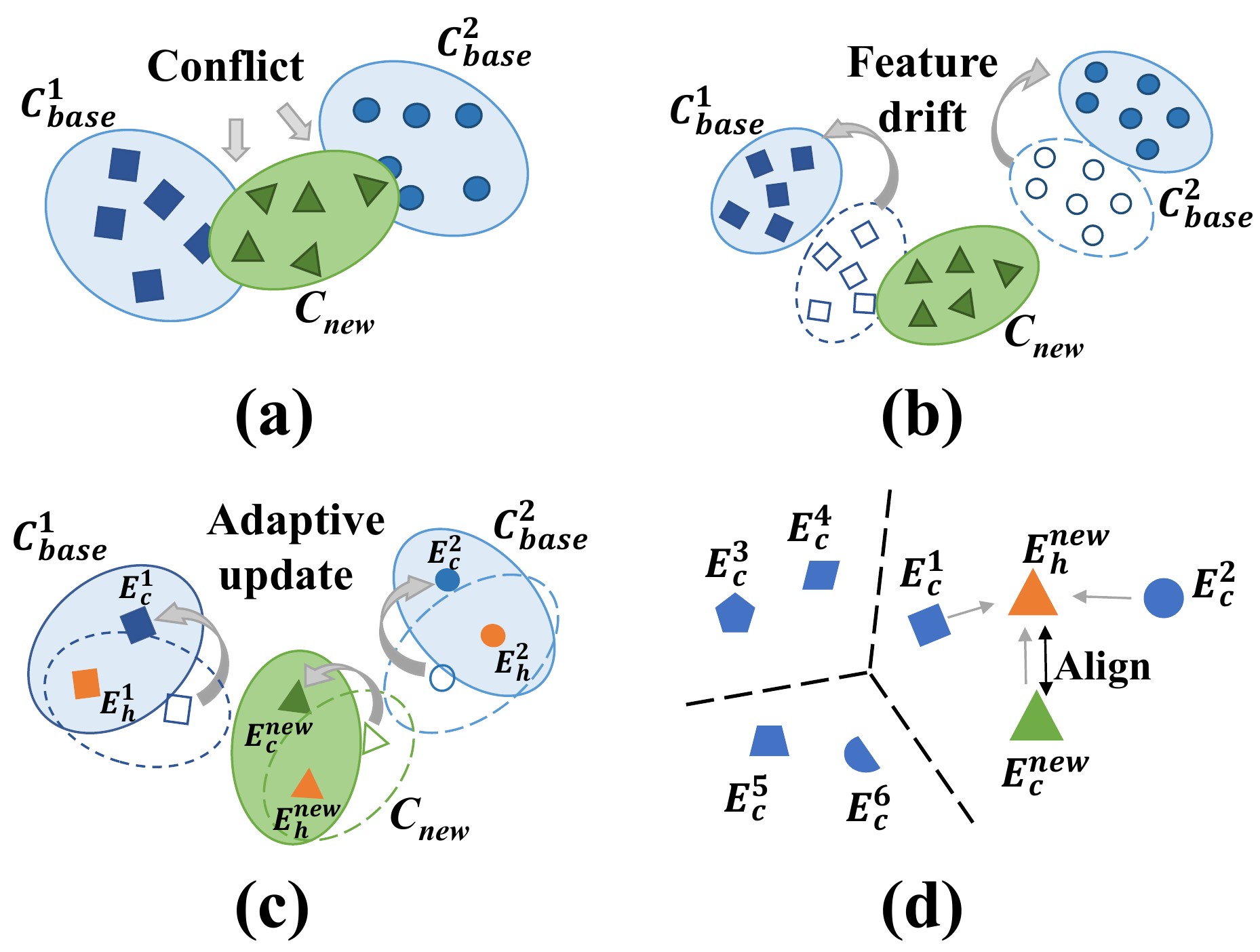}
  \caption{(a) In few-shot semantic segmentation, knowledge representation of base classes $C_{base}$ and new classes $C_{new}$ often conflicts, resulting in catastrophic forgetting problem. (b) Due to coupling between knowledge learning and representation, feature embeddings would drift to mis-match the true distribution of base classes. (c) To resist feature drift, EAUS maintains hyper-class embeddings $E_h$ to store old knowledge with representation and adaptively updates category embeddings $E_c$ to combine new class. (d) Hyper-class embedding $E^{new}_h$ is generated by firstly clustering category embeddings corresponding to base classes (in blue) and then aligning with category embedding of new class $E_c^{new}$.}
    \label{fig:F1}
    
\end{figure}

Few-shot semantic segmentation \cite{DBLP:conf/cvpr/LiJSSKK21,DBLP:conf/cvpr/LiuZLL20,DBLP:conf/icmcs/ShiWP0L21} addresses to segment a new category of images with few samples, decreasing the cost of expensive pixel-level annotations. In a real-world scenario, we expect the trained model to segment new classes without forgetting knowledge learned from old classes, which is a natural task for human beings. However, fine-tuning a deployed model with few samples of new classes leads to a severe catastrophic forgetting problem \cite{MCCLOSKEY1989109}, since models tend to forget knowledge about old classes when facing representation conflict between old and new classes as shown in Fig. \ref{fig:F1}(a). The gap between humans and machine learning models inspires researchers to facilitate incremental few-shot segmentation (IFSS), which aims to learn a segmentation model for both old and new classes with only few new samples.

The main challenges in IFSS are catastrophic forgetting of already acquired knowledge and overfitting networks to few samples of new classes.
Most current incremental methods \cite{8107520, 9578675, DBLP:conf/cvpr/ZhangSLZPX21, DBLP:conf/cvpr/GaneaBP21} use parameters or embedding vectors to represent category knowledge from a cognitive-inspired perspective, which addresses catastrophic forgetting by knowledge representation with new-class updating scheme. However, representation error for old classes would accumulate iteratively, since they couple knowledge learning and representation in each updating iteration as shown in Fig. \ref{fig:F1}(b), thus inevitably hindering to maintain useful and consistent knowledge learned from old classes.

In this paper, we propose EHNet, i.e., Embedding adaptive-update and Hyper-class representation Network for IFSS, addressing problems of catastrophic forgetting and overfitting.
We learn two kinds of embedding vectors, i.e., category and hyper-class embedding, from few samples of a new class, where the former describes exclusive semantical properties, and the latter expresses hyper-class knowledge as class-shared semantic properties.
One key benefit of such knowledge representation is to replace requirement of new parameters training with prediction on fixed-length semantic embeddings, thus preventing training from scratch when learning new classes.

To mitigate catastrophic forgetting, we propose an Embedding Adaptive-Update Strategy (EAUS) as shown in Fig. \ref{fig:F1}(c), where category embeddings are adaptively adjusted with an attention scheme and hyper-class embeddings remain unchanged.
In this manner, a well-separated representation of classes is built, where old knowledge is well maintained in memory functional design, i.e., hyper-class. EAUS decouples knowledge learning and representation to solve feature drift via selectively update, and alleviates the requirement of new-class sample number by keeping old knowledge. The core of category embeddings updating is a class-attention scheme, which computes a safe displacement vector for each class by contextualizing individual class weights over the representations of all classes. This adapted class-attention scheme not only highlights the discriminative representations between base and new classes to generate better decision boundaries over all involved classes, but also indicates directions towards a well-separated representation with less semantic biases during incremental learning sessions.

Observation for semantical segmentation proves to segment unseen classes with knowledge, where newly discovered samples may share semantic properties like 'haired' and 'quadruped' with the classes that have been learned.
The hyper-class is thus formulated as an abstract representation that contains semantic properties of similar classes, which enables to reduce data-scale dependence and overfitting by sharing semantic knowledge during learning sessions.
A hyper-class embedding is learned by clustering category embeddings of all classes for initialization and aligning with the new-class category embedding for enhancement, which is shown in Fig. \ref{fig:F1}(d).
On the one hand, the clustering algorithm is applied to the set of all category embeddings to generate a raw hyper-class embedding, thus extracting a similar semantic representation as new hyper-class knowledge.
On the other hand, we align the generated hyper-class embedding with new-class category embedding to enhance correlated semantic information and eliminate uncorrelated information. 

Significantly, EAUS builds a well-separated representation with few semantic biases. Meanwhile, the hyper-class knowledge complements and enhances semantic information.
These two designs provide representations for class with sufficient semantics but limited biases, thus enabling to well perform image segmentation tasks requiring high semantic dependence. 

In summary, the contributions of this paper are:
 
\begin{itemize}

  \item
  We propose an embedding adaptive-update strategy to avoid catastrophic forgetting, where hyper-class embeddings remain fixed to maintain old knowledge, and category embeddings are adaptively updated with a class-attention scheme, combining new classes learned in incremental sessions.

  \item
  To resist overfitting caused by few training samples, a hyper-class is firstly learned by clustering category embeddings and then aligned with new-class category embedding for correlation enhancement, thus alleviating performance dependence on training data scale.

  \item
  Experimental results show EHNet achieves state-of-the-art performance with remarkable advantages.

\end{itemize}

\section{Related work}

\subsection{Semantic Segmentation}

Semantic segmentation aims to classify each pixel of an image into a set of preset categories.
According to different network structures, current methods can be roughly divided into three categories, i.e., CNN-based, RNN-based and GNN-based methods.
CNN-based methods \cite{DBLP:conf/cvpr/LongSD15, DBLP:conf/cvpr/0005DSZWTA18, DBLP:conf/cvpr/LiSCLZWS20} utilize convolution operations to extract semantic information from feature maps for pixel-level label prediction.
Considering dependence of context information, RNN-based methods \cite{DBLP:conf/cvpr/VisinRCMCKBC16, DBLP:conf/miccai/MilletariRBEN18, DBLP:journals/bspc/KangZHHMI22} use recurrent layers to capture local and global spatial structure information of images.
Using topological structure of graphs, GNN-based methods \cite{DBLP:journals/corr/abs-2001-00335, DBLP:journals/corr/abs-2103-06509, DBLP:conf/miccai/MengWGZYHZ20} transform task of image segmentation into the classification task of graph nodes.

Among them, CNN-based methods have received more popularity.
For instance, to reconstruct high-resolution prediction images, UNet~\cite{DBLP:conf/miccai/RonnebergerFB15} and its variant~\cite{DBLP:journals/pami/BadrinarayananK17} use an encoder-decoder structure to segment images.
The encoder   downsamples the feature map to obtain a large field of view and capture abstract feature representations, while the decoder gradually restores fine-grained information.
Considering information loss caused by pooling, Chen et al. \cite{DBLP:journals/pami/ChenPKMY18} propose DeepLab, where dilated convolution is used to enlarge receptive fields while maintaining the resolution of feature maps.

\subsection{Few-Shot learning}

Few-shot learning aims to learn a model, which can be easily transferred to new tasks with limited training data.
We roughly divide current few-shot learning methods into two categories, i.e, initialization based and metric learning based methods.
The former methods \cite{DBLP:conf/icml/FinnAL17, DBLP:conf/iclr/RaviL17, DBLP:conf/cvpr/LeeMRS19, DBLP:conf/iclr/ChenLKWH19} generally define few-shot learning as "learning to fine-tune", which aims to learn proper model initialization or predict network parameters.
For example, MAML \cite{DBLP:conf/icml/FinnAL17} explicitly trains the parameters of model to produce good generalization performance, which is easy to perform another task with few training samples and gradient updating steps.
To get quick convergence within a few updates, Ravi et al. \cite{DBLP:conf/iclr/RaviL17} propose a LSTM-based meta-learner model with general initialization, specially designing for a few-shot learning scenario.

Metric learning based methods \cite{DBLP:conf/aaai/GongYB20, DBLP:conf/nips/VinyalsBLKW16, DBLP:conf/nips/SnellSZ17} firstly learn a projection function that projects the inputs to an embedding space, and then define a certain distance metric that measures the distance between any two embeddings.
For example, Siamese Network \cite{DBLP:conf/icml/SungYZXTH18} compares samples in query set and support set by calculating similarity between their extracted feature vectors, thus performing few-shot classification based on known category labels.

Facing the problem of domain shifts between training and test datasets, Yuan et al. \cite{DBLP:journals/pr/YuanCLWXZ22} propose a novel forget-update module,
 which could improve the discrimination by learning to forget and generate new features based on each task.
Different from current few-shot methods, EHNet utilizes old knowledge to help learn new knowledge, where new classes could share the semantic features of base classes, thus alleviating the dependence on data scale when learning new classes.

\subsection{Incremental learning}

Incremental learning \cite{DBLP:journals/corr/abs-2109-11369, DBLP:conf/iccv/LiKRHD021} studies how to extend the knowledge of a model without a performance drop on old knowledge. Previous works can be roughly grouped in two categories \cite{DBLP:journals/corr/abs-1909-08383}, i.e., replay-based and regularization-based methods.

In replay-based methods, samples of previous tasks are either stored or generated at first and then replayed when learning the new task. For example, Li et al. \cite{DBLP:journals/pami/LiH18a} propose to jointly learn new labels and base classes from the outputs of a pre-trained teacher model. Since it's not capable to properly distinguish between old and new classes, later approaches \cite{DBLP:conf/cvpr/LeeLSL19, DBLP:conf/eccv/CastroMGSA18} utilize the memory of old classes for further training by considering distribution relationship between base and new classes.

Regularization-based methods \cite{DBLP:conf/cvpr/DharSPWC19, DBLP:journals/pami/LiH18a, DBLP:conf/cvpr/GaneaBP21} protect old knowledge from being covered by imposing constraints on new tasks.
For example, to regularize the learning of new classes, Ren et al. \cite{DBLP:conf/nips/RenLFZ19} propose Attention Attractor Network, which utilizes old weights to train a set of new weights that could recognize new classes.
To make classifiers learned on individual sessions suitable for all classes, Zhang et al. \cite{DBLP:conf/cvpr/ZhangSLZPX21} propose a continually evolved classifier, which utilizes a graph model to propagate context information between classifiers for progressively adaptation.

Recently, incremental learning has been extensively studied for image segmentation task \cite{DBLP:journals/cviu/MichieliZ21, DBLP:conf/aaai/GuDW21, DBLP:conf/mm/YanZXZ021}. However, previous works in incremental segmentation focused on new classes that come with rich samples. In this paper, we explore the incremental segmentation task with few-shot setting.

\section{Task Description}

Incremental few-shot segmentation (IFSS) aims to generate a model that learns to segment new classes from few new samples without forgetting knowledge about old classes. IFSS has several learning sessions that come in sequence.
Once the learning of the model steps into the next session, the training datasets in previous sessions are no longer available. Meanwhile, evaluation in each session involves classes of all previous sessions and the current session.

Specifically, let \{$D^0_s,  D^1_s, \cdots, D^n_s$\} denote the support sets of different learning sessions, and the corresponding label space of dataset $D^i_s$ is denoted by $C^i$. Different datasets have no overlapped classes, i.e. $\forall i, j$ and $i{\neq}j$, $C^i{\cap}C^j={\emptyset}$.
At the $i$-th learning session, only $D^i_s$ can be used for network training.
For evaluation, the query set $D^i_q$ at session $i$ includes test data from all previous and current classes, i.e., the label space is $C^0 \cup C^1 \cdots \cup C^i$.
Usually, the support set $D^0_s$ in the first session is a large dataset, where sufficient samples are available for training.
On the contrary, support sets in all following sessions only include few training samples.

\section{Methodology}

\subsection{Overview}
\label{method:overview}

\begin{figure*}[h]
  \centering
  \includegraphics[width=\linewidth]{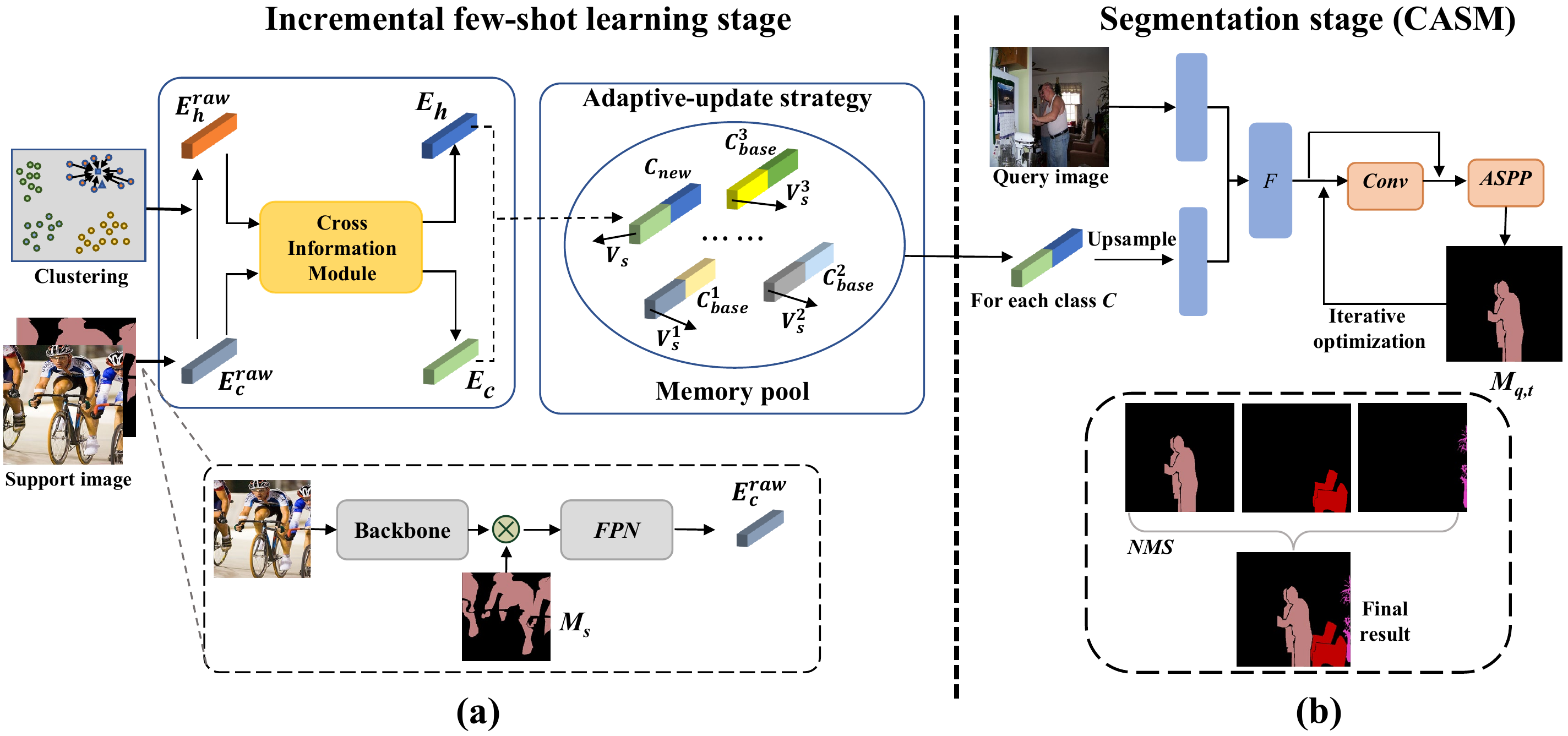}
  \caption{The design of EHNet. (a) In incremental few-shot learning stage, a support image is represented as a hyper-class embedding $E_h$ and a category embedding $E_c$. $E_h$ and $E_c$ are stored in memory pool, where category embeddings of all classes are adaptively updated to obtain a well-separated representation. (b) In segmentation stage, objects of each class are segmented, based on corresponding $E_h$ and $E_c$. Those segmentation results are integrated by non-maximum suppression (NMS) to generate the final result.}
    \label{architecture}
\end{figure*}

Facing problems of catastrophic forgetting and overfitting in IFSS, we propose EHNet, mainly including representation of hyper-class knowledge and an Embedding Adaptive-Update Strategy (EAUS). Hyper-class knowledge could provide additional semantic information, which alleviates the dependence on data scale during learning new classes, thus avoiding overfitting caused by few training samples. EAUS keeps hyper-classes fixed as memory to maintain old knowledge, and adaptively updates category embeddings with a class-attention scheme to obtain a well-separated representation with few semantic biases.
EHNet can be divided into two stages, i.e., incremental few-shot learning stage on support sets and segmentation stage on query sets, as shown in Figure \ref{architecture}.

In incremental few-shot learning stage, we first learn two embeddings, i.e., raw category embedding $E^{raw}_c$ and raw hyper-class embedding $E^{raw}_h$, from a training sample. 
$E^{raw}_c$ describes exclusive semantical properties, and $E^{raw}_h$ expresses hyper-class knowledge
as class-shared semantic properties.
Since $E^{raw}_h$ is obtained by clustering on base classes, there are semantic biases between $E^{raw}_c$ and $E^{raw}_h$.
To eliminate the semantic biases, $E^{raw}_c$ and $E^{raw}_h$ are semantically aligned through Cross Information Module (CIM), thus generating a category embedding $E_c$ and a hyper-class embedding $E_h$.
Finally, $E_c$ and $E_h$ are stored in memory pool, saving semantic embeddings to keep memory of the learned classes, where category embeddings of both base and new classes are updated via EAUS to obtain a well-separated representation with few semantic biases.

In segmentation stage, to match the aforementioned approach that keeps memory of base classes by saving embeddings, we propose Class-Agnostic Segmentation Module (CASM) as shown in Figure. \ref{architecture} (b), which segments each class based on the corresponding semantic embeddings in memory pool.

\subsection{Semantic Embedding}
\label{method:embedding}

To reduce the overfitting issues caused by few training samples, we use category embedding and hyper-class embedding to jointly describe feature representation of a class.
The prediction network has not seen a new class, while the new class may share semantic properties with base classes it has seen.
For instance, the network has never seen tigers before. However, many typical attributes of tigers can be found from base classes (e.g., cat, leopard, lion). We use hyper-class embedding to represent the similar attributes of base classes, thus assisting the network to understand tigers by reducing the dependence on training data scale. 

The generation of category embedding and hyper-class embedding is shown in Figure \ref{architecture} (a).
First, a support image is input into backbone network to generate feature maps. Then, the feature maps are multiplied by a binary mask $M_s$ to remove irrelevant background, which outputs feature representation only containing target objects.
Afterwards, the feature representation is input into Feature Pyramid Network (FPN) \cite{DBLP:conf/cvpr/LinDGHHB17}, which extracts high-level semantic information and generates a raw category embedding $E_c^{raw}$.
Hereafter, we perform clustering on base classes and obtain a raw hyper-class embedding $E_h^{raw}$ based on $E_c^{raw}$.
Finally, Cross Information Module (CIM) semantically aligns $E_h^{raw}$ and $E_c^{raw}$ to generate a hyper-class embedding $E_h$ and a category embedding $E_c$.
which could enhance category-relevant information and eliminate irrelevant information in hyper-class embeddings.

In $k$-shot learning, embeddings of new classes are updated via our EAUS during learning multiple samples.
Owing to multiple updates, features of the $k$ samples are fused, which generates an enhanced embedding representation with more semantic information. 
Details of fusing multiple samples are described in Section \ref{method:update}.

\textbf{Category Embedding.}
Category embeddings describe exclusive semantical property of classes.
Considering variances of target objects in size, we use feature pyramid network (FPN) \cite{DBLP:conf/cvpr/LinDGHHB17} to extract semantic information in different levels. By global pooling, the output of FPN is compressed into a 512-dimensional feature vector as a raw category embedding.

\textbf{Hyper-class Embedding.}
Hyper-class embedding expresses hyper-class knowledge as shared semantic properties, which is generated by clustering on base classes and aligning semantic information with the corresponding new-class category embedding. Specifically, for a new class, our network searches for similar base classes through $K$-means clustering, which computes the center of all category embeddings as a raw hyper-class embedding for the new class.
The raw hyper-class embedding is then aligned in semantic information with the new-class category embedding, thus enhancing correlated semantic information and eliminating irrelevant information.
Owing to hyper-class embedding, new classes can share the semantic information of base classes, where old knowledge could help learn new knowledge, thus reducing the dependence on training data scale.

It's noted the number of selected similar base classes has an impact on hyper-class knowledge. A small number of similar classes can't guarantee to provide sufficient shared semantic information, meanwhile a large number of similar classes could bring noise by introducing irrelevant classes.  

\textbf{Cross Information Module.}
Since hyper-class embeddings generated by a untrainable clustering algorithm may not semantically match the new-class category embedding, we propose cross information module (CIM) to achieve semantic alignment between hyper-class and category embedding, thus enhancing category-relevant information and eliminating irrelevant information in hyper-class embeddings.

\begin{figure}[t]
  \centering
  \includegraphics[width=\linewidth]{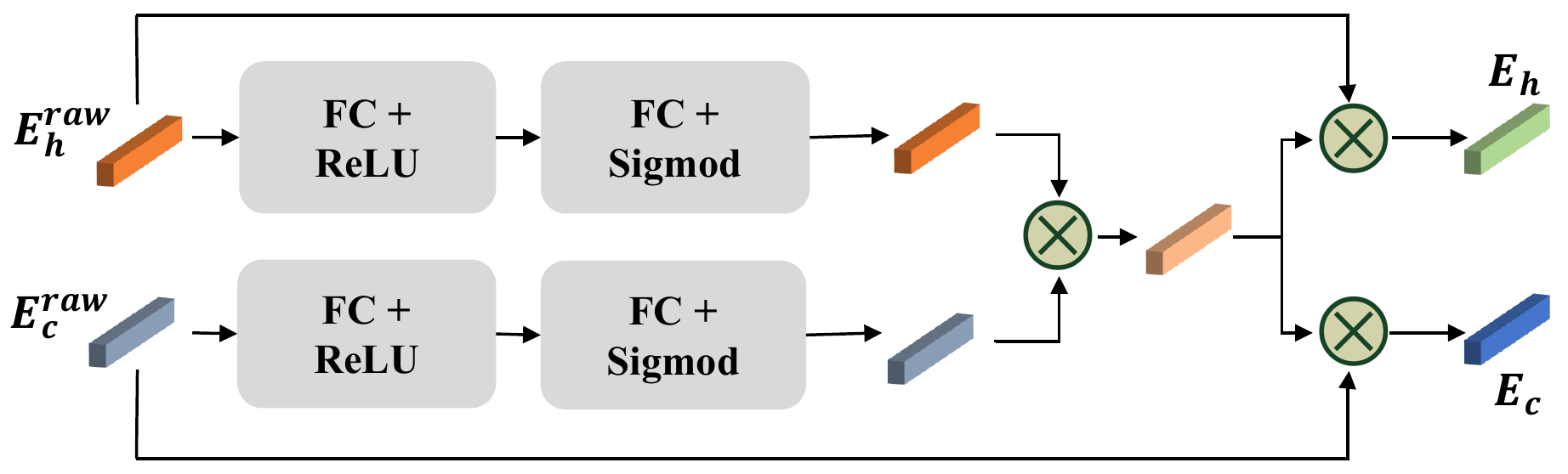}
  \caption{The design of CIM. Given raw hyper-class embedding $E^{raw}_a$ and raw category embedding $E^{raw}_c$, CIM generates updated category embedding $E_c$ and hyper-class embedding $E_h$, which achieves semantic alignment between these two embeddings.}
  \label{CIM}
\end{figure}

The design of CIM is shown in Figure \ref{CIM}.
Firstly, raw hyper-class embedding $E_h^{raw}$ and raw category embedding $E_c^{raw}$ are sent to a pair of two-layer fully-connected (FC) layers, respectively. The Sigmoid activation function attached after the FC layer transforms vector values into importance weights of channels.
Afterwards, the embedding vectors in two branches are fused by element-wise multiplication.
Intuitively, only similar semantic features  would own a high activation value in the fused vector.
Finally, we adopt the fused vector to weight the raw embedding vectors, thus generating enhanced embedding representations.
In comparison to the raw embeddings, the enhanced embeddings focus on the correlated semantic information, thus obtaining semantic alignment between the new-class hyper-class and category embeddings.


\subsection{Embedding Adaptive-Update Strategy}
\label{method:update}

To resist catastrophical forgetting, we design Embedding Adaptive-Update Strategy (EAUS), which selectively and adaptively updates embeddings of base and new classes, thus mitigating feature drift and generating well-separated embedding representation for all classes.

Specifically, we keep hyper-class embeddings consistent as memory to maintain old knowledge in learning sessions, thus reducing feature drift effect.
First, we compute a relation coefficient $e_{i,j}$ between class $i$ and class $j$ based on their category embeddings $E_c^i$ and $E_c^j$:

\begin{equation}
  e_{i,j} = \langle \varPhi(E_c^i),\varPsi(E_c^j) \rangle
\end{equation}
where $\varPhi(\cdot)$ and $\varPsi(\cdot)$ are linear transformation functions that project the original representations to a new metric space, and $\langle\cdot,\cdot\rangle$ is a similarity function that computes the inner product between two embedding vectors.

Then, we normalize all the coefficients with a softmax function to obtain the attention weight $a_{i,j}$ of two classes:

\begin{equation}
  a_{i,j} = softmax(e_{i,j}) =  \frac{exp(e_{i,j})}{\sum_{l = 1}^{|P|} exp(e_{i,l}) }
\end{equation}
where $|P|$ represents the number of classes in memory pool.

Afterwards, we perform subtraction operation on the embedding vectors of class $i$ and $j$ to obtain a subtraction vector, which locally indicates the updating direction.
Finally, Based on the normalized attention weight and the corresponding subtraction vector, we calculate a safe displacement vector $V_s$ for each category embedding. Considering information of all classes, $V_s$ 
indicates directions towards a well-separated representation with less semantic biases
during  learning sessions. Therefore, the category embedding $E_c^i$ can be adaptively updated as $E_c^i{'}$ by guidance of $V_s$:

\begin{equation}
  E_c^i{'} = E_c^i +  \sum_{l = 1}^{|P|} a_{j,l}W(E_c^i-E_c^l)
\end{equation}
where $W(\cdot)$ is a linear transformation. It's noted that we repeat the operations above to update category embeddings of all classes. 

In $k$-shot learning, we directly apply embeddings of new samples to update the learned embeddings in EAUS, thus fusing semantic information from multiple samples without an additional fusion module.
Specifically, for a class $i$, we define subtraction vectors in $k$-shot learning as $E_c^i-E_c^j$ with different class $j$, and $E_c^l-E_c^i$ with the same class $l$. 
It's noted that we force $E_c^i-E_c^j$  and $E_c^l-E_c^i$ to separate and fuse embeddings, thus forming well-separated feature representation and enhancement of same-class semantic information, respectively. 
In fact, embeddings of the same class generally obtain a larger relation coefficient due to similar values, thus achieving an effective fusion effect in EAUS.

\subsection{Class-Agnostic Segmentation Module}
\label{method:segmentation}

Class-Agnostic Segmentation Module (CASM) segments each class based on corresponding embeddings in memory pool, without considering what class an embedding represents. 
The design of CASM is shown in Figure \ref{architecture} (b).
First, a query image is processed by backbone network to generate feature maps, which contain target objects of different classes.
For each target class, we perform an up-sampling operation on the semantic embeddings, and concatenate them with the feature maps to obtain feature maps $F$ for dense comparison. After concatenation, we process feature maps via a 3*3 convolution block with a residual connection.
Afterwards, we adopt an atrous spatial pyramid pooling module proposed in \cite{DBLP:journals/pami/ChenPKMY18} to capture multi-scaled information and output the segmentation result, which is iteratively optimized to obtain compact boundary predictions. The process can be formulated as:

\begin{equation}
  M_{q,t+1} = f_A(F\oplus conv(F+M_{q,t}))
\end{equation}
where $\oplus$ represents element-wise addition, + represents concatenation operation, function $conv(\cdot)$ refers to convolution operation, and $f_A(\cdot)$ represents the process of atrous spatial pyramid pooling module.  $M_{q,t}$ and $M_{q,t+1}$ refer to the predicted masks from current iteration step and the next iteration step, respectively.

Finally, for each pixel $p$, class with the highest confidence is selected as output class label by non-maximum suppression (NMS) algorithm:

\begin{equation}
 NMS(p) = \arg\max\limits_{c\in \mathcal{C}}{H(p,c)}
\end{equation}
where $\mathcal{C}$ represents the set of all learned classes, and $H(p,c)$ represents the probability that the pixel $p$ belongs to the class $c$.

It's noted that backbone networks in learning stage and segmentation stage are the same in structure and parameters, since it's necessary to represent support and query images in the same feature space for few biases. Specifically, we use the first four layers of Resnet-50 \cite{DBLP:conf/cvpr/HeZRS16} pre-trained on base dataset $D^0_s$ as backbone network.

\section{Experiments}

\subsection{Dataset and Metric}

\textit{PASCAL-5$^i$} \cite{DBLP:conf/bmvc/ShabanBLEB17} is a widely used few-shot segmentation dataset, which consists of PASCAL VOC 2012 \cite{DBLP:journals/ijcv/EveringhamGWWZ10} and additional annotations in SDS \cite{DBLP:conf/eccv/HariharanAGM14}.
The dataset includes 20 categories, which are divided into 4 splits, and each split contains 5 categories.

\textit{COCO 2014} \cite{DBLP:conf/eccv/LinMBHPRDZ14} is a challenging large-scale dataset containing 80 categories.
 COCO is designed for natural scene understanding by acquiring data from complex daily scenes, where
 targets in images are segmented by precise pixel-level labels.
The original dataset contains 82783 training images and 40504 validation images.

Based on the task description of incremental few-shot segmentation, we make special settings on datasets.
On PASCAL-5$^i$, the first split is considered as a base dataset, each class thus containing sufficient samples. Other splits are used for incremental learning in different sessions, each class only containing few samples for new-class training.
On COCO 2014, we divide the 80 classes into 4 splits, and each split contains 20 classes. Similarly, the first split is a base dataset, and other splits are incremental few-shot datasets.

Following \cite{DBLP:conf/bmvc/ShabanBLEB17}, we measure the per-class foreground Intersection-over-Union (IoU) and use the mean IoU over all classes (mIoU) to report the results.

\subsection{Implementation Details}

We implement EHNet using Pytorch library, and train it for 200 epochs on four Nvidia 1080Ti GPUs.
We adopt cross-entropy loss function to evaluate the segmentation loss, and use StepLR scheduler in training, which reduces the learning rate (initial value as 0.0001)  to 0.9 times for every 20 epochs of training. We evaluate performance of EHNet by setting $k=1$ and $5$ shots per new class. To ensure reliable results with random sample selection in $k$-shots learning, we run all tests 10 times and report the mean result for comparisons.

\subsection{Comparisons with Other Methods}

Since there exists few incremental few-shot segmentation methods for comparisons, 
we modify current few-shot segmentation and incremental few-shot learning methods for IFSS task.
For few-shot segmentation methods, i.e., CANet \cite{DBLP:conf/cvpr/ZhangLLYS19} and ARNet \cite{DBLP:conf/icmcs/ShiWP0L21}, we save embeddings of classes for new-class segmentation, meanwhile we either non-update or apply EAUS to update embeddings for comparisons. 
For incremental few-shot learning method, i.e., AAN \cite{DBLP:conf/nips/RenLFZ19} and XtarNet \cite{DBLP:conf/icml/YoonKSM20}, we apply CASM to them for goal of performing segmentation. Besides, we compare EHNet with several incremental segmentation methods, e.g., MiB \cite{DBLP:conf/cvpr/CermelliMB0C20} and EMNet \cite{DBLP:conf/mm/YanZXZ021}.

\begin{table*}
  \centering
  \caption{Comparison results of 1-shot and 5-shot segmentation on PASCAL-5$^i$ dataset.  Best in bold.}
  \label{comparion1}
  \begin{tabular}{c|ccccccc|ccccccc}
    \hline
  \multirow{3}{*}{Method} & \multicolumn{7}{c|}{1-shot}                               & \multicolumn{7}{c}{5-shot}                                                                                 \\
                          & session-0 & \multicolumn{2}{c}{session-1} & \multicolumn{2}{c}{session-2} & \multicolumn{2}{c|}{session-3} & session-0 & \multicolumn{2}{c}{session-1} & \multicolumn{2}{c}{session-2} & \multicolumn{2}{c}{session-3}  \\
                          & New     & Base & New                  & Base & New                  & Base & New                  & New     & Base & New                  & Base & New                  & Base & New                   \\ \hline
  MiB \cite{DBLP:conf/cvpr/CermelliMB0C20}         & 48.3      & 25.4 & 27.2                   & 19.4 & 20.3                   & 16.2 & 12.9                   & 52.7      & 30.8 & 31.3                   & 29.6 & 25.7                   & 21.6 & 19.6                    \\
  EMNet \cite{DBLP:conf/mm/YanZXZ021}              & 50.8      & 21.8 & 18.0                   & 16.8 & 13.5                   & 10.7 & 10.9                   & 53.5      & 36.3 & 32.3                   & 25.4 & 24.9                   & 19.6 & 15.8                    \\
  CANet \cite{DBLP:conf/cvpr/ZhangLLYS19} +INC     & 51.5      & 33.1 & 38.8                   & 26.2 & 28.4                   & 19.5 & 23.9                   & 55.5      & 37.2 & 39.6                   & 31.3 & 34.6                   & 21.6 & 25.2                    \\
  CANet \cite{DBLP:conf/cvpr/ZhangLLYS19} +EAUS      & 52.1      & 46.6 & 43.1                   & 37.2 & 39.6                   & 32.6 & 34.6                   & 54.5      & 49.3 & 46.3                   & 39.4 & 40.9                   & 35.6 & 36.4                    \\
  ARNet \cite{DBLP:conf/icmcs/ShiWP0L21} +INC       & 54.0      & 33.4 & 40.4                   & 27.7 & 31.3                   & 22.6 & 25.8                   & 55.9      & 40.3 & 38.8                   & 30.9 & 33.1                   & 24.4 & 22.1                    \\
  ARNet \cite{DBLP:conf/icmcs/ShiWP0L21} +EAUS       & 54.8      & 47.9 & 48.2                   & 41.1 & 42.9                   & 37.0 & 36.9                   & 56.4      & 49.0 & 47.2                   & 43.1 & 43.8                   & 38.5 & 40.2                    \\
  AAN \cite{DBLP:conf/nips/RenLFZ19} +CASM           & 48.2      & 44.6 & 42.6                   & 38.5 & 35.8                   & 33.8 & 34.2                   & 49.1      & 45.3 & 44.3                   & 39.4 & 37.6                   & 34.6 & 35.5                    \\
  XtarNet \cite{DBLP:conf/icml/YoonKSM20} +CASM     & 47.4      & 43.7 & 41.8                   & 38.7 & 37.2                   & 34.7 & 32.0                   & 50.7      & 45.1 & 44.9                   & 40.9 & 41.2                   & 37.8 & 36.8                    \\
  PIFS \cite{DBLP:conf/bmvc/CermelliMXAC21}          & 53.9      & 48.1 & 46.2                   & 43.6 & 41.2                   & 38.2 & 37.4                   & 54.2      & 49.2 & 47.5                   & 43.9 & 42.6                   & 40.1 & 39.4                    \\  
  EHNet (ours)      & \textbf{56.7} & \textbf{51.4} & \textbf{53.2}  & \textbf{46.3} & \textbf{46.8}  & \textbf{40.9} & \textbf{41.8}   & \textbf{57.1}   & \textbf{53.4} & \textbf{55.2} & \textbf{50.5} & \textbf{51.2}  & \textbf{44.6} & \textbf{45.7}         \\ \hline
  \end{tabular}
  \end{table*}

\begin{table}
  \centering
  \caption{Comparison results of 1-shot and 5-shot segmentation on COCO dataset. Best in bold.}
  \label{comparion2}
  \begin{tabular}{c|cc|cc}
    \hline
  \multirow{2}{*}{Method}                     & \multicolumn{2}{c|}{1-shot}       & \multicolumn{2}{c}{5-shot}                                                                                 \\
                                                   & Base       & New             & Base    & New     \\ \hline
  MiB \cite{DBLP:conf/cvpr/CermelliMB0C20}         & 15.1       & 17.0              & 18.4    & 19.4      \\
  EMNet \cite{DBLP:conf/mm/YanZXZ021}              & 10.7	      & 11.8  	          & 15.8	  & 18.9    \\
  CANet \cite{DBLP:conf/cvpr/ZhangLLYS19} +INC    & 15.6       & 13.2	            & 19.7	  & 14.9      \\
  CANet \cite{DBLP:conf/cvpr/ZhangLLYS19} +EAUS    & 23.3       &	28.1              &	24.8    &	30.1      \\
  ARNet \cite{DBLP:conf/icmcs/ShiWP0L21} +INC      & 13.7       &	11.6              &	10.6    & 11.4      \\
  ARNet \cite{DBLP:conf/icmcs/ShiWP0L21} +EAUS     & 22.8       &	26.1              &	27.7    &	30.1      \\
  AAN \cite{DBLP:conf/nips/RenLFZ19} +CASM         & 27.2       & 29.5              & 28.4    &	29.7      \\
  XtarNet \cite{DBLP:conf/icml/YoonKSM20} +CASM    & 26.1       & 28.6              &	25.7    & 26.8      \\
  PIFS \cite{DBLP:conf/bmvc/CermelliMXAC21}        & 28.8       & 31.4              &	29.7    & 31.8      \\
  EHNet (ours)                      & \textbf{29.7}    & \textbf{33.1}       & \textbf{33.4}    & \textbf{36.6}  \\ \hline
\end{tabular}
\end{table}

\textbf{\textit{PASCAL-5$^i$.}}
Comparison results on PASCAL-5$^i$ are shown in Table \ref{comparion1}, where ``INC'' means the embeddings of classes are saved for incremental learning and keep unchanged without updating. Table \ref{comparion1} shows EHNet is superior to other methods, achieving new state-of-the-art performance.  

As the number of learned classes increases, all methods face a great challenge in learning new classes without forgetting base classes. 
It's noted that the performance of few-shot segmentation methods (CANet\cite{DBLP:conf/cvpr/ZhangLLYS19} and ARNet\cite{DBLP:conf/icmcs/ShiWP0L21}) without updating drops significantly, due to the conflicts between feature representations of new and base classes. 
Equipped with EAUS, few-shot segmentation methods greatly improve their performance by comparing performance without updating, which proves that EAUS significantly mitigates catastrophic forgetting by maintaining hyper-class embeddings as old knowledge and adaptively updating category embeddings as source of new-class knowledge.  The incremental segmentation methods (MiB\cite{DBLP:conf/cvpr/CermelliMB0C20} and EMNet\cite{DBLP:conf/mm/YanZXZ021}) perform poorly in few-shot settings, since they generally require large number of samples for new-class learning.

The improvement in segmentation performance with different learning sessions proves that incremental few-shot learning methods (AAN\cite{DBLP:conf/nips/RenLFZ19} and XtarNet\cite{DBLP:conf/icml/YoonKSM20}) could reduce catastrophic forgetting by iteratively learning new-class knowledge. However, due to feature drift in knowledge representation and non-separated embeddings with insufficient semantics,
they have limited ability in segmentation task that requires high semantics for accurate pixel-level labeling. 
PIFS couples prototype learning and knowledge distillation for IFSS, 
while its representation error for old classes would accumulate iteratively,
thus inevitably hindering to maintain useful and consistent knowledge learned from old classes.
 
Moreover, comparison methods fail to model commonalities between old and new classes. In other words, they can't utilize old knowledge to help better learn new-class embedded knowledge, which is proved by poor segmentation performance on new classes. On the contrary, EHNet introduces hyper-class embeddings as old knowledge to share semantic properties of base classes with new-class learning, which alleviates the dependence on training samples scale of new classes and thereby avoids overfitting issues.

\textbf{\textit{COCO.}}
Comparison results on COCO dataset are shown in Table \ref{comparion2}, where EHNet achieves the best performance on COCO dataset as well.  
Regarding that PASCAL-5$^i$ only contains 20 categories and COCO contains 80 categories, increasing in category number requires a high distinguish capability of models to compute pixel-level predictions, which can be proved by general decrease in performance when comparing results on PASCAL-5$^i$ and COCO achieved by the same method. 
As shown in Table \ref{comparion2}, performance of those methods without strategies to update embeddings drops significantly when learning session increases. Such phenomenon can be explained by the fact that more classes involved in COCO result in more obvious conflicts in embedding representation, thus causing a significant catastrophic forgetting effect. 
Meanwhile, experimental results show that our EAUS can effectively mitigate catastrophic forgetting by a smaller drop in performance.

\begin{figure*}[h]
  \centering
  \includegraphics[width=\linewidth]{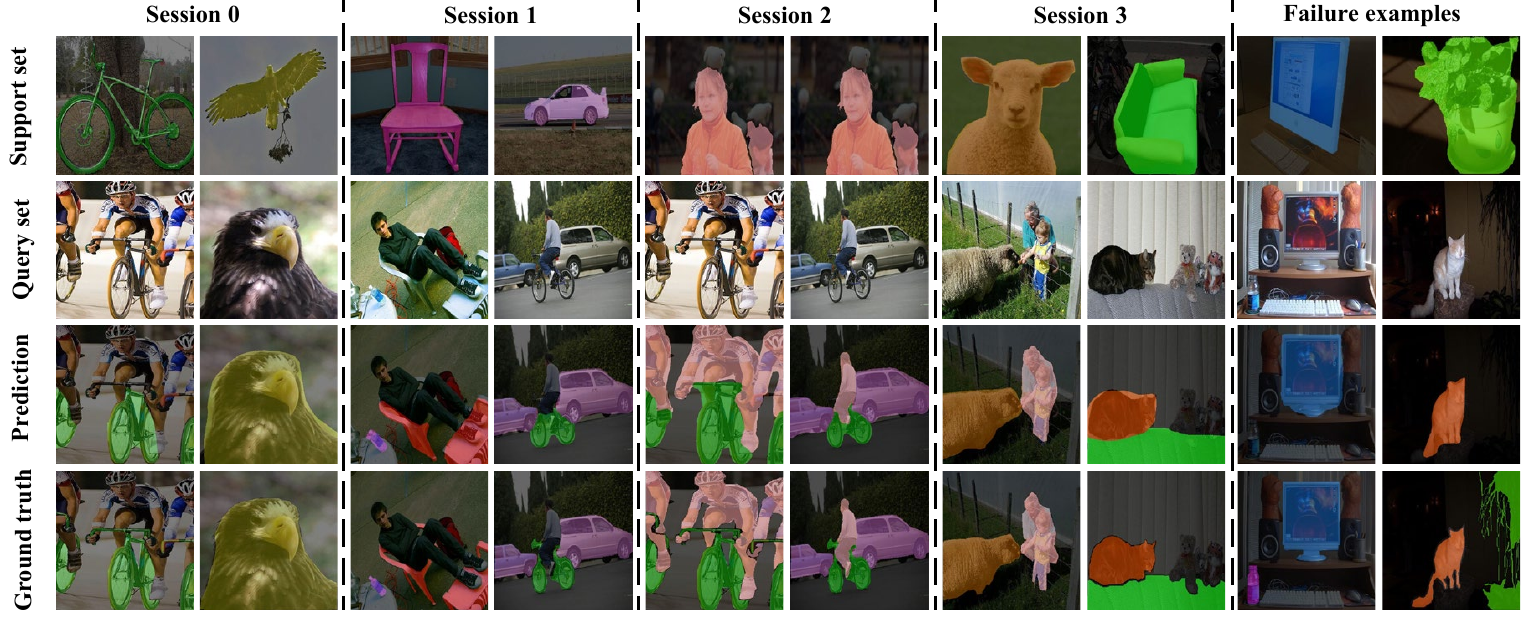}
  \caption{Qualitative results of EHNet, where we show the segmentation results in different sessions and some failure examples. The new classes in the current session will become base classes in subsequent sessions.}
    \label{visualization}
\end{figure*}

\textbf{\textit{Qualitative results.}}
Some qualitative results of 1-shot segmentation are shown in Figure \ref{visualization}, where each row represents the support set, query set, prediction, and ground-truth, respectively.
From left to right, we show the segmentation results with four different sessions and some failure cases, respectively. 
With the increase of sessions, more classes are required to be segmented by EHNet, which brings difficulties in segmentation with new-class learning. 
From the failure examples, we can observe that segmentation of small objects or in dim environments is still challenging.

\subsection{Ablation Study}

To prove effectiveness of the proposed modules, we implement ablation experiments on PASCAL-5$^i$ dataset to report average mIOU performance of 4 splits with 1-shot setting.

\begin{table}[t]
  \centering
  \caption{Results achieved with different settings on semantic embeddings. Best in bold.}
  \label{embedding}
  \begin{tabular}{cc|ccc}
    \hline
  $E_h$ & $E_c$ & Base & New & Mean   \\  \hline
  $\surd$  & $\times$  & 22.7 & 25.3  & 24.0     \\
  $\times$  & $\surd$  & 35.3 & 30.6  & 33.0  \\
  $\surd$  & $\surd$  & 41.6 & 46.4  & 44.0    \\
  $\surd$  & $\bigcirc$  & 45.4 & 44.9  & 45.2  \\
  $\bigcirc$   & $\surd$  & \textbf{46.2} & \textbf{49.6}  & \textbf{47.9}    \\  \hline
  \end{tabular}
 
  \end{table}

\textbf{\textit{Semantic embedding.}}
To validate the effectiveness of constructing hyper-class embeddings ($E_h$), category embeddings ($E_c$), and the strategy to keep $E_h$ and update $E_c$ in EAUS, we compare experiment results by either removing or updating certain embeddings. Results are shown in Table \ref{embedding}, where ``$\times$'', ``$\surd$'' and ``$\bigcirc$'' represent the embedding is removed, updated, and unchanged, respectively.

Table \ref{embedding} shows the elimination of certain embeddings would reduce the performance of EHNet, which means that each semantic embedding plays a positive role in segmentation.
When both embeddings are updated, segmentation performance on base classes drops significantly, which proves that updating both embeddings leads to semantic feature drift and poor performance.

\textbf{\textit{Embedding adaptively-update strategy.}}
To validate the effectiveness of updating strategy in EAUS, we apply different embedding update strategies for comparisons, e.g., non-update strategy and trainable linear transformation (LT) strategy. To explore whether embeddings of base classes should be updated in EAUS, we adaptively update embeddings of either base class ($C_{base}$) or new class ($C_{new}$).

Results are shown in Table \ref{class}, where ``$\surd$'' presents corresponding embeddings are updated, and ``Non-update'' presents only storing embeddings without updates.
Table \ref{class} shows EHNet without embedding updating is unsatisfied in performance, indicating catastrophic forgetting problem occurs in incremental learning without intervention. Meanwhile, LT alleviates forgetting problem to a certain extent proved by relatively better performance, but it's still limited in building well-separated representation. 
When EAUS is only applied to base or new classes, performance is degraded, indicating that globally modeling of embedding space for incremental learning is hardly achieved by only updating base or new classes.

\begin{table}
  \centering
  \caption{Results with different strategies on updating embeddings. Best in bold.}
  \label{class}
  \begin{tabular}{ccc|ccc}
    \hline
  Method      & $C_{base}$   & $C_{new}$ & Base & New & Mean   \\  \hline
  Non-update  & $\surd$   & $\surd$  & 15.6 & 18.7  & 17.2     \\
  LT          & $\surd$  &      & 24.1 & 23.7  & 23.5  \\
  LT          &     & $\surd$  & 20.2 & 25.6  & 22.9     \\
  LT          & $\surd$   & $\surd$  & 25.2 & 27.3  & 26.3  \\
  EAUS (ours)   & $\surd$   &         & 44.1 & 34.8  & 39.5     \\
  EAUS (ours)   &     & $\surd$       & 36.5 & 47.9  & 42.2  \\
  EAUS (ours)    & $\surd$  & $\surd$  & \textbf{46.2} & \textbf{49.6}  & \textbf{47.9}    \\  \hline
  \end{tabular}

\end{table}

\textbf{\textit{Number of iterations in CASM.}}
To validate the effectiveness of iterative optimization in CASM, we compare segmentation results and speed with different iteration numbers in Table \ref{Iterations}.
It's noted the segmentation speed is measured by the number of processed frames per second (FPS).
Table \ref{Iterations} shows segmentation performance gets better and speed gets slower with the increasing number of iterations.
Being larger than 4 iterations, raising the number of iterations doesn't contribute to segmentation performance, which proves that 4 iterations in CASM keep a balance between performance and computation cost. Essentially, excessive iterations would lead our model to favor obvious objects with clear boundaries, thus harming segmentation of objects with less salience in query images.

\begin{table}
  \centering 
  \caption{Results of EHNet with different numbers of iterations. Best in bold.}
  \label{Iterations}
  \begin{tabular}{c|cccc}
    \hline
  Iterations  & Speed  & Base & New & Mean   \\  \hline
  0           & \textbf{24.7}   & 42.7 & 45.7  & 44.2     \\
  1           & 16.4   & 44.1 & 46.9  & 45.5     \\
  2           & 12.4   & 45.0 & 48.6  & 46.8     \\
  3           & 9.8    & 45.7 & 49.5  & 47.6     \\
  4           & 8.2    & 46.2  & \textbf{49.6}  & \textbf{47.9}     \\ 
  5           & 6.9    & \textbf{46.4} & 48.6  & 47.5     \\
  \hline
  \end{tabular}
  
  \end{table}

\section{Conclusion}

In this paper, we focus on IFSS task and present EHNet.
To avoid catastrophic forgetting, we propose an embedding adaptive-update strategy, where hyper-class embedding keeps unchanged as memory to maintain old knowledge, and category embeddings are adaptively updated to combine new classes learned on incremental sessions.
To resist overfitting when learning with few samples of new classes, we learn hyper-class embeddings by clustering and aligning with category embeddings, where new classes could share semantic features of base classes, thus alleviating the dependence on training data scale.
Comprehensive experiments show that EHNet achieves new state-of-the-art performance.

\section*{Acknowledgments}
This work was supported by Key Laboratory of Water Big Data Technology of Ministry of Water Resources, in part by a grant from National Key R\&D Program of China under Grant No. 2021YFB3900601, 
the Fundamental Research Funds for the Central Universities under Grant B220202074,
and National Research Foundation, Singapore underits AI Singapore Programme (AISG Award No: AISG-100E-2020-065).

\bibliographystyle{ACM-Reference-Format}
\bibliography{reference}


\end{document}